\documentclass[runningheads]{llncs}
\usepackage{verbatim}
\usepackage{amsmath}
\usepackage{amssymb}
\usepackage{bm}
\usepackage{booktabs} 
\usepackage{multirow}
\usepackage{multicol} 
\usepackage{graphicx}
\usepackage{url}
\usepackage{amsfonts}
\usepackage{graphicx}
\usepackage{wrapfig}
\usepackage{arydshln}
\usepackage{color}
\usepackage{bbding}
\usepackage{hyperref}
\usepackage{hyperref}
\begin{document}
%
% \title{Contribution Title\thanks{Supported by organization x.}}
\title{Consisaug: A Consistency-based Augmentation for Polyp Detection in Endoscopy Image Analysis}
\titlerunning{Consisaug: A Consistency-based Augmentation for Polyp Detection}
\author{Ziyu Zhou\inst{1} \and Wenyuan Shen\inst{2} \and Chang Liu\inst{3} \Envelope}
\authorrunning{Z. Zhou et al.}
\institute{$^1$ Shanghai Jiao Tong University, Shanghai, China \\
{\tt zhouziyu@sjtu.edu.cn}\\
$^2$ Carnegie Mellon University, PA, USA \\
{\tt wenyuan2@andrew.cmu.edu}\\
$^3$ SenseTime Research, Shanghai, China \\
{\tt liuchang@sensetime.com}}

\maketitle              % typeset the header of the contribution
\begin{abstract}
Colorectal cancer (CRC), which frequently originates from initially benign polyps, remains a significant contributor to global cancer-related mortality. Early and accurate detection of these polyps via colono-scopy is crucial for CRC prevention. However, traditional colonoscopy methods depend heavily on the operator's experience, leading to suboptimal polyp detection rates. Besides, the public database are limited in polyp size and shape diversity. To enhance the available data for polyp detection, we introduce Consisaug, an innovative and effective methodology to augment data  that leverages deep learning. We utilize the constraint that when the image is flipped the class label should be equal and the bonding boxes should be consistent. We implement our Consisaug on five public polyp datasets and at three backbones, and the results show the effectiveness of our method. All the codes are available at (\href{https://github.com/Zhouziyuya/Consisaug}{https://github.com/Zhouziyuya/Consisaug}).

\end{abstract}
\section{Introduction}

Colonoscopy, while essential for colorectal cancer (CRC) screening, is expensive, resource-demanding, and often met with patient reluctance. Unfortunately, up to 26\% of colonoscopies may miss lesions and adenomas~\cite{zhao2019magnitude}, as they heavily rely on the expertise of the endoscopist. In routine examinations, distinguishing between neoplastic and non-neoplastic polyps poses challenges, especially for less experienced endoscopists using current equipment~\cite{wadhwa2020physician}~\cite{dayyeh2015asge}. Besides, object detection labeling involves the expertise of the endoscopist assigning both a category and a bounding box location to each object in an image. This process is time-consuming, with an average of 10 seconds per object~\cite{russakovsky2015best}. Consequently, object detection labeling incurs significant costs, demands extensive time commitments, and requires substantial effort.

Recently,  there has been a great interest in deep learning in CRC screening. Various studies have developed models for automatic polyp segmentation~\cite{tomar2022transresu}~\cite{fan2020pranet}, polyp detection~\cite{sun2022maf}~\cite{jiang2023yona} aiming to reduce the access barrier to pathological services. However, the deficiency of training data seriously impedes the development of polyp detection techniques. The existing fully-annotated databases, including CVC-ClinicDB\cite{bernal2015wm}, ETIS-Larib\cite{silva2014toward}, CVC-ColonDB\cite{bernal2012towards}, Kvasir-Seg\cite{borgli2020hyperkvasir} and LDPolypVideo\cite{ma2021ldpolypvideo}, are very limited in polyp size and shape diversity, which are far from the significant complexity in the actual clinical situation. Therefore, in this paper we want to find out an augmentation to fully use the dataset itself. By this motivation, we put forward an consistency-based augmentation to improve the performance of polyp detection which use Student-Teacher model to distill knowledge. Following we coarsely describe the consistency regularization and Student-Teacher model used in our architecture.

Consistency regularization is a method that has seen wide applications in semi-supervised learning, unsupervised learning, and self-supervised learning. The core idea behind consistency regularization is to encourage the model to produce similar outputs for similar inputs, thereby leveraging unlabeled data to improve generalization performance\cite{laine2016temporal}\cite{tarvainen2017mean}\cite{miyato2018virtual}. Student-Teacher models have been a focal point of research in the field of machine learning and specifically in the domain of knowledge distillation, where information is transferred from one machine learning model (the teacher) to another (the student). The objective is to leverage the capabilities of a large, complex model (the teacher) and distill this knowledge into a smaller, simpler model (the student), thereby optimizing computational efficiency without compromising the performance significantly~\cite{hinton2015distilling}~\cite{xie2020self}.

Through this work, we have made the following contributions:
\begin{itemize}
    \item[$\bullet$] We propose a straightforward yet effective augmentation scheme that take advantage of the polyp image's intrinsic flipping consistency property;
    \item[$\bullet$] We novelly combine flipping consistency with Student-Teacher architecture which show great effectiveness in polyp detection;
    \item[$\bullet$] The proposed consistency constraint augmentation for polyp detection works well on multiple datasets and backbones and effective for not only in-domain samples but cross-domain samples.
\end{itemize}

\section{Method}

\begin{figure}[h!]
    \centering
    \setlength{\belowcaptionskip}{0.1cm}
    \includegraphics[width=12cm]{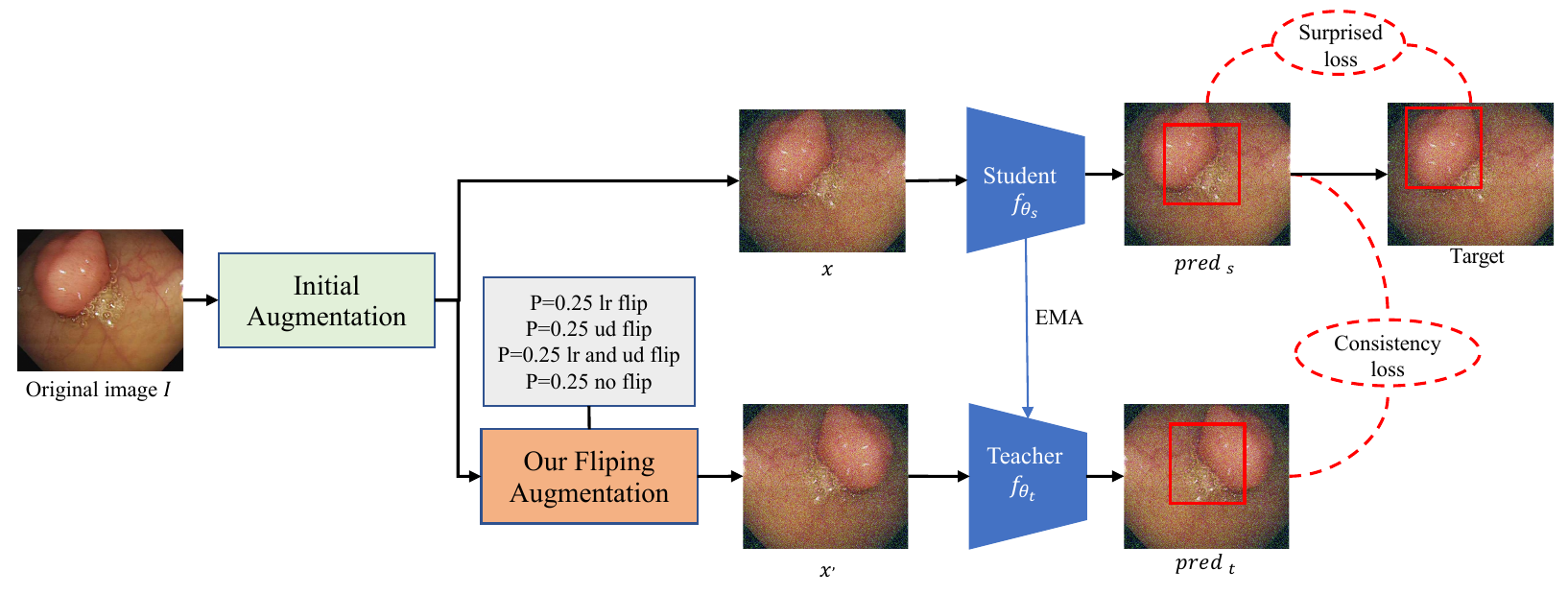}
    \caption{Overall structure of our proposed method.
}
    \label{fig:pipeline}
\end{figure}

The Consisaug to be presented works similarly depending on whether it is for a CNN-based or a transformer-based object detector. The overall structure is depicted in Fig. \ref{fig:pipeline}. The proposed structure is the combination of the Student-Teacher model and an object detection algorithm. To allow one-to-one correspondence of target objects, an original image $I$ is added to initial augmentation to get image $x$. And $x$ is added to our flipping augmentations to get the flip one $x^\prime$. As shown in Fig. \ref{fig:pipeline}, a paired bounding box should represent the same class and their localization information should be consistent.\\
\\In the Student branch, the labeled samples are trained using supervised loss in typical object detection approaches. The consistency loss is additionally used to combine the two outputs of the teacher and student model. In this section, the Student-Teacher model, consistency loss for localization and for classification will be introduced respectively.

\subsection{The Student-Teacher Model}
\label{student-teacher}

The framework of Consisaug used for this work shares the same overall structure as recent knowledge distillation approaches. There are two parts of the model \emph{Student} and \emph{Teacher} shown in Fig. \ref{fig:pipeline}. The \emph{Student} model outputs predictions $f(I)$ of input images: $f(I) \triangleq f_{\theta_s}(I)$ and the same as \emph{Teacher} model: $f(\hat{I}) \triangleq f_{\theta_t}(\hat{I})$, $f(I)$ and $f(\hat{I})$ are a paired of bounding boxes which should represent the same class and consistent localization, $I$ and $\hat{I}\in {R}^{C \times H \times W}$ are the two input images of the models after respective augmentations. Differently, in the student branch, the original image $I_0$ will be randomly added initial augmentation including addnoise, multi-scale, flip and etc to get image $I$, while in the teacher branch, $I$ will be randomly added our flipping augmentations. The two models have the same architecture and the same initial weight. Besides, the weights of \emph{Student} $\theta_s$ are updated by back-propagation and the weights of \emph{Teacher} $\theta_t$ are updated by exponential moving average as Eq. \ref{eq:ema}. More precisely, the temperature $\tau \in [0, 1]$ is given and updated after each iteration.
\begin{equation}\label{eq:ema} 
    \theta_t \leftarrow \tau \theta_s+(1-\tau)\theta_t
\end{equation}

% \subsection{Consistency loss for classification}

% We denote $f_{cls}^{k}(I)$ as the output class probability vector after softmax operation corresponding to the prediction box. Since $\hat{I}$ is a flipping augmented version of $I$, the predictions of the two images should be equivalent. In unsupervised learning, some candidates such as $L_2$ distance or Jensen-Shannon divergence (JSD) can be used as the consistency regularization loss. Among them, we specifically take advantage of JSD for the following reasons. 

\subsection{Consistency loss for localization}

We denote $f_{loc}^{k}(I)$ as the output localization of the rediction box. The localization result for the $k$-th candidate box $f_{loc}^k(I)$ consists of $[\Delta c x, \Delta c y, \Delta w, \Delta h]$, which represent the displacement of the center and scale coefficient of a candidate box, respectively. The output $f_{loc}^k(I)$ and its flipping version $f_{loc}^k(\hat{I})$ require a simple modification to be equivalent to each other. Since our flipping transformations make the coordinate offset move into the opposite direction, a negation should be applied to correct them. And we take the left and right flipping as an example in Eq. \ref{flip_example}
\begin{equation} \label{flip_example}
\begin{aligned} 
\Delta c x^k & \Longleftrightarrow-\Delta c \hat{x}^{k^{\prime}} \\
\Delta c y^k, \Delta w^k, \Delta h^k & \Longleftrightarrow \Delta c y y^{k^{\prime}}, \Delta \hat{w}^{k^{\prime}}, \Delta \hat{h}^{k^{\prime}}
\end{aligned}
\end{equation}
The localization consistency loss used for a single pair of bounding boxes in our method is given as below:
\begin{equation}
\begin{aligned}
l_{con\_loc }(f_{l o c}^k(I), f_{l o c}^{k^{\prime}}(\hat{I}))= & \frac{1}{4}\left(\left\|\Delta c x^k-\left(-\Delta \hat{c x} \hat{x}^{k^{\prime}}\right)\right\|^2+\left\|\Delta c y^k-\Delta \hat{c y} \hat{k}^{k^{\prime}}\right\|^2\right. \\
& \left.+\left\|\Delta w^k-\Delta \hat{w}^{k^{\prime}}\right\|^2+\left\|\Delta h^k-\Delta \hat{h}^{k^{\prime}}\right\|^2\right)
\end{aligned}
\end{equation}
The overall consistency loss for localization is then obtained from the average of loss values from all bounding box pairs:
\begin{equation}
\mathcal{L}_{con-l }=\mathbb{E}_k\left[l_{con\_loc }\left(f_{l o c}^k(I), f_{l o c}^{k^{\prime}}(\hat{I})\right)\right]
\end{equation}

\subsection{Consistency loss for classification}

As for consistency loss for classification, we use Jensen-Shannon divergence (JSD) instead of the $L_2$ distance as the consistency regularization loss. $L_2$ distance treats all the classes equal, while in our flipping consistency circumstance irrelevant classes with low probability should not effect the classification performance much. JSD is a weaker constraint loss which is suitable in the consistency setting. The classification consistency loss is defined as below:

\begin{equation}
l_{c o n \_c l s}\left(f_{c l s}^k(I), f_{c l s}^{k^{\prime}}(\hat{I})\right)=J S\left(f_{c l s}^k(I), f_{c l s}^{k^{\prime}}(\hat{I})\right)
\end{equation}
where $JS$ denotes Jensen-Shannon divergence and $f_{c l s}^k(I)$ is the model prediction class of the $k$-th box in image $I$. The overall consistency loss for classification of a pair of flipping images can be clarified as:

\begin{equation}
\mathcal{L}_{con-c}=\mathbb{E}_k\left[l_{con\_cls}\left(f_{cls}^k(I), f_{cls}^{k^{\prime}}(\hat{I})\right)\right]
\end{equation}

The total consistency loss is the sum of location and classification consistency loss:

\begin{equation}
\mathcal{L}_{con}=\mathcal{L}_{con-l}+\mathcal{L}_{con-c}
\end{equation}
Consequently, the final loss $\mathcal{L}$ is composed of the original object detector's fully supervised loss $\mathcal{L}_{s}$ and our consistency loss $\mathcal{L}_{con}$:

\begin{equation}
\mathcal{L}=\mathcal{L}_{s}+\mathcal{L}_{con}
\end{equation}

% While the object detection is only polyp in our setting, the classification item contributes no ingredient to the total loss.

\section{Experiments and Results}

\subsection{Implementation Details}

\textbf{Datasets and Baselines.} Experiments are conducted on 5 public polyp datasets LDPolypVideo\cite{ma2021ldpolypvideo}, CVC-ColonDB\cite{bernal2012towards}, CVC-ClinicDB\cite{bernal2015wm}, Kvasir-Seg\cite{borgli2020hyperkvasir} and ETIS-Larib\cite{silva2014toward}. The first one is the largest-scale challenging colonoscopy polyp detection dataset and the others are standard benchmarks for polyp segmentation. We summarize the 5 datasets by listing their parameters in Table \ref{tab:dataset}. We use the official train test split on LDPolypVideo dataset and split 10\% data from train set for validation. As for datasets with no officially released partition, we split 80\% for training and 10\% for validating and testing respectively.

We train our Consisaug on three baseline models to evaluate the effectiveness of our method: yolov5\cite{jocher2022ultralytics}, SSD\cite{liu2016ssd} and detr\cite{carion2020end}. The first two are CNN-based models and the third one is Transformer-based model. All experiments have been done under the similar setting of the official codes and both codes are implemented based on Pytorch. The evaluation metrics used in our experiments are recall, precision, mAP50, F1-score, F2-score and the last three is vital in image detection. In detail, we use AdamW~\cite{loshchilov2017decoupled} optimizer with a cosine learning rate schedule, linear warm up of 10 epochs while the overall epoch is 100, and 0.0001 for the maximum learning rate value. The batch size is 32 and the image size is 640. We train with single Nvidia RTX3090 24G GPU to proceed each experiment.

\begin{table}[hbp]
        \centering
        \setlength{\belowcaptionskip}{0.2cm}
        \caption{Summary of public annotated colonoscopy datasets.}
        \label{tab:dataset}
        \resizebox{0.75\columnwidth}{!}{
        % \scalebox{0.8}{
        \begin{tabular}{cccccc}
        \toprule    
             %\multirow{2}*{Backbone} & Pretraining data & Method & \multirow{2}*{ChestX-ray14} & \multirow{2}*{CheXpert} & \multirow{2}*{ShenZhen} & RSNA  \\
             % & and methods& & & & & Pneumonia \\
             Dataset &Label& Resolution & $N_{images}$ & $N_{videos}$ & $N_{polyps}$ \\
        \midrule
            LDPolypVideo & Bounding box& $560\times480$ & 33884 & 160 & 200 \\
            CVC-ClinicDB & Mask & $384\times288$ & 612 & 29 & 29\\
            CVC-ColonDB & Mask & $574\times500$ & 380 & 15 & 15\\
            ETIS-Larib& Mask & $1225\times966$ & 196 & 34 & 44\\
            Kvasir-Seg & Mask & Various & 1000 & N/A & N/A\\

        \bottomrule
        \end{tabular}
        }
    \end{table}

\subsection{Results} \label{sec:results}

\noindent \textbf{Consisaug outperforms the vanilla version on different backbones.} To demonstrate the effectiveness of our method, we train the polyp detection on three backbones yolov5\cite{jocher2022ultralytics}, SSD\cite{liu2016ssd}, and DETR\cite{carion2020end}, and all are trained on the LDPolypVideo dataset\cite{ma2021ldpolypvideo}, which has the largest size and diversity among the publicly released polyp datasets. The vanilla version model is trained using the official code and hyper-parameters, while the Consisaug version is trained using our method, which is reconstructed with the student-teacher model and our consistency-based augmentation. The results are shown in Table \ref{tab:differ-model}. All methods are trained three times, and the best results for each baseline are bolded. From the results, we can conclude that our Consisaug can enhance the polyp detection not only on CNN-based backbones (yolov5, SSD) but transformer-based backbone (DETR) from the three evaluation indexes mAP50, F1-score, and F2-score. Moreover, our method can also improve the recall, which is vital for lesion detection in medical image analysis.

\begin{table}[hbp]
% \tiny
        \centering
        \setlength{\belowcaptionskip}{0.2cm}
        \caption{The polyp detection results on LDPolypVideo dataset. The best results for each baseline are bolded.}
        \label{tab:differ-model}
        \resizebox{1.0\columnwidth}{!}{
        % \scalebox{0.8}{
        \begin{tabular}{cc|ccccc}
        \toprule    
             %\multirow{2}*{Backbone} & Pretraining data & Method & \multirow{2}*{ChestX-ray14} & \multirow{2}*{CheXpert} & \multirow{2}*{ShenZhen} & RSNA  \\
             % & and methods& & & & & Pneumonia \\
             Baseline&Method & Recall & Precision & mAP50 & F1-score & F2-score \\
        \midrule
            \multirow{2}*{yolov5} &Vanilla & 0.378$\pm$0.008 & \bf{0.578$\pm$0.012} & 0.510$\pm$0.017 & 0.457$\pm$0.010 & 0.406$\pm$0.007\\
            &Consisaug  & \bf{0.453$\pm$0.004} & 0.575$\pm$0.015 & \bf{0.540$\pm$0.024} & \bf{0.507$\pm$0.018} & \bf{0.473$\pm$0.011}\\
        \midrule
            \multirow{2}*{SSD} &Vanilla & 0.658$\pm$0.028 & 0.152$\pm$0.006 & 0.515$\pm$0.013 & 0.248$\pm$0.009& 0.396$\pm$0.011\\
            &Consisaug  & \bf{0.667$\pm$0.024} & \bf{0.155$\pm$0.003} & \bf{0.527$\pm$0.014} & \bf{0.251$\pm$0.010} & \bf{0.401$\pm$0.006}\\
        \midrule
            \multirow{2}*{DETR} &Vanilla & 0.584$\pm$0.026 & 0.446$\pm$0.013 & 0.468$\pm$0.011 & 0.506$\pm$0.017 & 0.550$\pm$0.016\\
            &Consisaug  & \bf{0.629$\pm$0.030} & \bf{0.480$\pm$0.013} & \bf{0.504$\pm$0.015} & \bf{0.544$\pm$0.016} & \bf{0.592$\pm$0.022}\\

        \bottomrule
        \end{tabular}
        }
\end{table}

\noindent \textbf{Consisaug shows effectiveness on different colonoscopy datasets.} To further verify the validity of our Consisaug method, we train the vanilla version and our Consisaug on other datasets. All the experiments are implemented on yolov5, and the results are shown in Table \ref{tab:differ-dataset}. Consisaug outperforms the vanilla version in at least four mAP50, F1-score, and F2-score metrics on five datasets.

\begin{table}[hbp]
% \tiny
        \centering
        \setlength{\belowcaptionskip}{0.2cm}
        \caption{The polyp detection results based on yolov5 baseline for different datasets. The best results for each dataset are bolded.}
        \label{tab:differ-dataset}
        \resizebox{1.0\columnwidth}{!}{
        % \scalebox{0.8}{
        \begin{tabular}{cc|ccccc}
        \toprule    
             %\multirow{2}*{Backbone} & Pretraining data & Method & \multirow{2}*{ChestX-ray14} & \multirow{2}*{CheXpert} & \multirow{2}*{ShenZhen} & RSNA  \\
             % & and methods& & & & & Pneumonia \\
             Dataset&Method & Precision & Recall & mAP50 & F1-score & F2-score \\
        \midrule
            \multirow{2}*{LDPolypVideo} &Vanilla & 0.378$\pm$0.008 & \bf{0.578$\pm$0.012} & 0.510$\pm$0.017 & 0.457$\pm$0.010 & 0.406$\pm$0.007\\
            &Consisaug  & \bf{0.453$\pm$0.004} & 0.575$\pm$0.015 & \bf{0.540$\pm$0.024} & \bf{0.507$\pm$0.018} & \bf{0.473$\pm$0.011}\\
        \midrule
            \multirow{2}*{CVC-ClinicDB} &Vanilla & 0.933$\pm$0.002 & 0.781$\pm$0.006 & 0.865$\pm$0.009 & 0.850$\pm$0.003 & 0.807$\pm$0.007\\
            &Consisaug  & \bf{0.967$\pm$0.002} & \bf{0.932$\pm$0.013} & \bf{0.963$\pm$0.004} & \bf{0.949$\pm$0.011} & \bf{0.939$\pm$0.015}\\
        \midrule
            \multirow{2}*{CVC-ClolonDB} &Vanilla & \bf{0.997$\pm$0.003} & 0.789$\pm$0.013 & 0.891$\pm$0.005 & 0.881$\pm$0.007& 0.823$\pm$0.001\\ 
            &Consisaug  & 0.970$\pm$0.001 & \bf{0.842$\pm$0.001} & \bf{0.916$\pm$0.002} & \bf{0.901$\pm$0.004}& \bf{0.865$\pm$0.001}\\
        \midrule
            \multirow{2}*{ETIS-Larib} &Vanilla & \bf{0.982$\pm$0.003} & 0.350$\pm$0.005 & \bf{0.634$\pm$0.009} & 0.516$\pm$0.007 & 0.402$\pm$0.003\\
            &Consisaug & 0.800$\pm$0.0005 & \bf{0.450$\pm$0.012} & 0.629$\pm$0.003 & \bf{0.576$\pm$0.002} & \bf{0.493$\pm$0.004}\\
        \midrule
            \multirow{2}*{Kvasir-Seg} &Vanilla & \bf{0.937$\pm$0.006} & 0.730$\pm$0.008 & 0.848$\pm$0.001 & \bf{0.821$\pm$0.007} & 0.764$\pm$0.002\\
            &Consisaug & 0.879$\pm$0.003 & \bf{0.762$\pm$0.002} & \bf{0.857$\pm$0.005} & 0.816$\pm$0.002 & \bf{0.783$\pm$0.006}\\

        \bottomrule
        \end{tabular}
        }
    \end{table}

\noindent \textbf{Consisaug transcends the vanilla version on cross-domain datasets.} We also validate our method on cross-domain colonoscopy datasets detection. The vanilla and Consisaug versions are all trained on LDPolypVideo dataset yolov5 backbone. We test the two checkpoints on the other four whole datasets and the results are shown in Table \ref{tab:cross-domain}. The results show the transferability of our model which is trained on one domain  and tested on the other domains. Our method's performance exceeds the vanilla version on all datasets from the metric of mAP50 and surpasses the vanilla version on three datasets from the F1-score and F2-score.

\begin{table}[hbp]
% \tiny
        \centering
        \setlength{\belowcaptionskip}{0.2cm}
        \caption{Cross-domain polyp detection results. The four different datasets are test using vanilla yolov5 model and our Consisaug yolov5 model. The two models are all trained on LDPolypVideo dataset. The best results for each dataset are bolded.}
        \label{tab:cross-domain}
        \resizebox{0.75\columnwidth}{!}{
        % \scalebox{0.8}{
        \begin{tabular}{cc|ccccc}
        \toprule    
             %\multirow{2}*{Backbone} & Pretraining data & Method & \multirow{2}*{ChestX-ray14} & \multirow{2}*{CheXpert} & \multirow{2}*{ShenZhen} & RSNA  \\
             % & and methods& & & & & Pneumonia \\
             Dataset&Method & Precision & Recall & mAP50 & F1-score & F2-score \\
        \midrule
            \multirow{2}*{CVC-ClinicDB} &Vanilla & \bf{0.783} & 0.598 & 0.716 & 0.678 & 0.628\\
            &Consisaug  & 0.782 & \bf{0.652} & \bf{0.746} & \bf{0.711} & \bf{0.674}\\
        \midrule
            \multirow{2}*{CVC-ClolonDB} &Vanilla & 0.780 & 0.578 & 0.701 & 0.664 & 0.610\\
            &Consisaug  & \bf{0.769} & \bf{0.639} & \bf{0.721} & \bf{0.698} & \bf{0.661}\\
        \midrule
            \multirow{2}*{ETIS-Larib} &Vanilla & 0.725 & \bf{0.625} & 0.713 & \bf{0.671} & \bf{0.643}\\
            &Consisaug & \bf{0.852} & 0.548 & \bf{0.719} & 0.667 & 0.590\\
        \midrule
            \multirow{2}*{Kvasir-Seg} &Vanilla & 0.706 & 0.638 & 0.707 & 0.670 & 0.651\\
            &Consisaug & \bf{0.755} & \bf{0.632} & \bf{0.736} & \bf{0.688} & \bf{0.653}\\

        \bottomrule
        \end{tabular}
        }
    \end{table}

% \smallskip

\begin{figure}[h!]
    \centering
    \setlength{\belowcaptionskip}{0.1cm}
    \includegraphics[width=12.2cm]{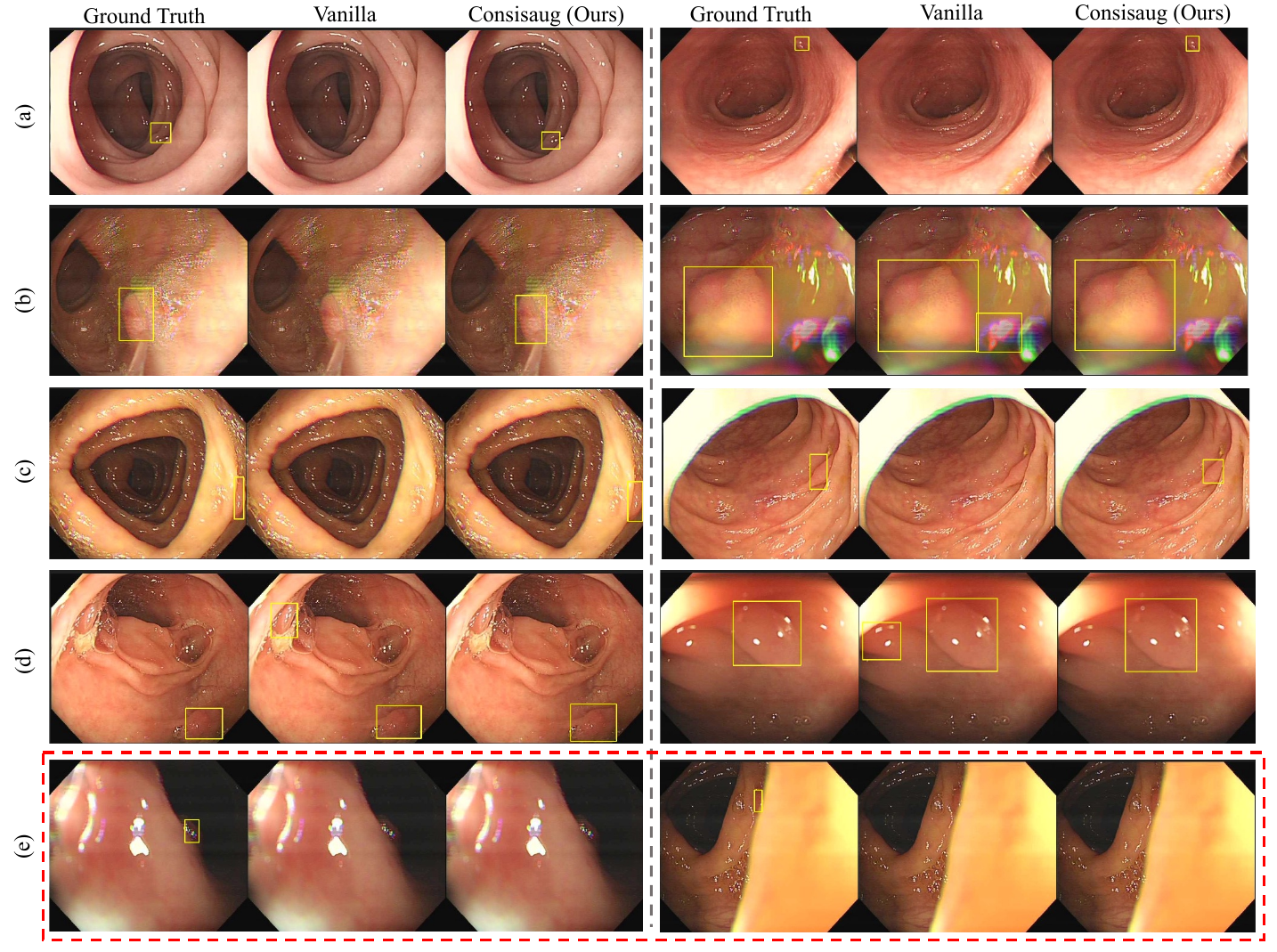}
    \caption{There are three columns for each image set. The first column is the image with ground truth, the second column shows the detection results of vanilla model and the third column is the results of our Consisaug method. The qualitative results prove that our Consisaug can (a) detect small targets, (b) detect targets in motion blur and reflections images, (c) detect targets between colon folds, (d) reduce false positive samples. And in (e) there will also be some failure cases for the hard detecting polyps.}
    \label{fig:visualization}
\end{figure}

\noindent \textbf{Qualitative Results.} In Fig. \ref{fig:visualization}, we provide the polyp detection results of our Consisaug on LDPolypVideo test set. Our method can locate the polyp tissues in many challenging cases, such as small targets, motion blur and reflection images, polyps between colon folds, etc. But there are also some failure cases detection for the low image quality or the targets hidden in the dark.

\section{Ablation Study}

\begin{table}[hbp]
% \tiny
        \centering
        \setlength{\belowcaptionskip}{0.2cm}
        \caption{The ablation study results on LDPolypVideo dataset yolov5 baseline. The best results for each baseline are bolded.}
        \label{tab:ablation}
        \resizebox{1.0\columnwidth}{!}{
        % \scalebox{0.8}{
        \begin{tabular}{ccc|ccccc}
        \toprule    
             Sup loss & Consisaug& Flip aug & Recall & Precision & mAP50 & F1-score & F2-score \\
        \midrule
            \Checkmark &\XSolidBrush &\XSolidBrush & 0.364$\pm$0.004 & 0.589$\pm$0.002 & 0.501$\pm$0.001 & 0.450$\pm$0.003 & 0.394$\pm$0.007\\
            \Checkmark&\XSolidBrush&\Checkmark  & 0.378$\pm$0.008 & 0.578$\pm$0.012 & 0.510$\pm$0.017 & 0.457$\pm$0.010 & 0.406$\pm$0.007\\
            \Checkmark& \Checkmark &\XSolidBrush & 0.386$\pm$0.008 & 0.554$\pm$0.003 & 0.515$\pm$0.003 & 0.455$\pm$0.009& 0.411$\pm$0.011\\
            \Checkmark& \Checkmark &\Checkmark & \bf{0.453$\pm$0.004} & \bf{0.575$\pm$0.015} & \bf{0.540$\pm$0.014} & \bf{0.507$\pm$0.018} & \bf{0.473$\pm$0.011}\\
        \bottomrule
        \end{tabular}
        }
\end{table}

\noindent In this section, we test the component of our Consisaug to provide deeper insight into our model. The supervised loss with flipping augmentations shown in Table \ref{tab:ablation} is the vanilla version used in section \ref{sec:results}. The ablation studies can be split into four combinations: (a) the model only uses supervised loss; (b) the model uses supervised loss and flipping augmentations; (c) the model uses supervised loss combining with our Consisaug and (d) the model uses supervised loss, flipping augmentations and Consisaug. Comparing (b) and (c) we can infer that our flipping consistency augmentation Consisaug is more effective than the pure flipping augmentations. And from the results in (d), combining our Consisaug with flipping augmentations can get the best performance which further proves the validity of our method.

\section{Conclusion}

We propose Consisaug, a novel Student-Teacher based augmentation for lesion detection task. Our approach takes advantage of the characteristics of colonoscopic surgery, in which the lens can be rotated at any angle in the body so the flip of the colonoscopy picture at any angle is the image state that can be obtained. Therefore, we leverage the peculiarity of the colonscopies and the flip detecting consistency to prove our method. Extensive experiments demonstrate that Consisaug is a valid augmentation across five datasets and three backbones. 

\bibliographystyle{unsrt}
\bibliography{egbib}
\end{document}